# Detectiona and Classification of Acute Lymphoblastic Leukemia Utilizing Deep Transfer Learing

Md. Abu Ahnaf Mollick
*Dept. Of Computer Science and Engineering*
*Varendra University*
Rajshahi, Bangladesh
mollickavoy@gmail.com

Md. Mahfujur Rahman
*Dept. Of Computer Science and Engineering*
*Varendra University*
Rajshahi, Bangladesh
mahfujur@vu.edu.bd

D.M. Asadujjaman
*Dept. Of Computer Science and Engineering*
*Varendra University*
Rajshahi, Bangladesh
asadujjaman2207557@stud.kuet.ac.bd

Abdullah Tamim
*Dept. Of Computer Science and Engineering*
*Varendra University*
Rajshahi, Bangladesh
tamim.cse.vu@gmail.com

Nosin Anjum Dristi
*Dept. Of Computer Science and Engineering*
*Varendra University*
Rajshahi, Bangladesh
nosindristi040@gmail.com

Md. Takbir Hossen
*Dept. Of Computer Science and Engineering*
*Varendra University*
Rajshahi, Bangladesh
tahsinahmedtakbir@gmail.com

*Abstract*— A mutation in the DNA of a single cell that compromises its function initiates leukemia. This leads to the overproduction of immature white blood cells, which encroach upon the space required for the generation of healthy blood cells. Leukemia is treatable if identified in its initial stages. Nonetheless, its diagnosis is both arduous and time-consuming. In this study, a novel approach for diagnosing leukemia across four stages—Benign, Early, Pre, and Pro—utilizing deep learning techniques. We employed two Convolutional Neural Network (CNN) models: MobileNetV2 with an altered head and a bespoke model. The custom model has multiple convolutional layers, each paired with corresponding max pooling layers. We utilized MobileNetV2 with ImageNet weights, and the head was adjusted to integrate the final results. The utilized dataset is a publicly available collection of blood cell smear images titled "Acute Lymphoblastic Leukemia (ALL) image dataset", and then used the Synthetic Minority Oversampling Technique (SMOTE) to augment and balance the training dataset. Which attained an accuracy of 98.6% with the custom model, while MobileNetV2 achieved a superior accuracy of 99.69%. The pre-trained model exhibited encouraging results and an increased likelihood of real-world application.

*Keywords— Leukemia, MobileNetV2, CNN, SMOTE*

## I. Introduction

The identification of healthy blood cells is essential in the diagnosis of leukemia. Leukemia is etymologically derived from the Greek terms "leukos," signifying "white," and "aim," meaning "blood." There are generally four prevalent types of leukemia, namely acute lymphoblastic leukemia (ALL), chronic lymphocytic leukemia (CLL), acute myeloid leukemia (AML), and chronic myeloid leukemia (CML) [1]. Leukemia is characterized by the development of malignant leukocytes within the bone marrow. As a result, the typical defense mechanism of the human body is compromised. White blood cells constitute a fundamental component of the human immune system [2]. Leukemia is a severe cancer common in South Asia. In 2020, South Asia reported over 62,000 leukemia cases, with India having the highest and Bangladesh the lowest incidence. In 2020, leukemia ranked as the 15th most diagnosed cancer and the 11th leading cause of cancer-related mortality worldwide [3]. Leukemia necessitates early detection for accurate diagnosis, raising concern. Leukemia detection is labor-intensive, necessitating trained personnel to interpret blood cell images via microscopes [4]. Bangladesh is a developing country with a limited quantity of trained personnel [3], which is a matter of concern as leukemia requires early detection for proper diagnosis.

Deep learning is a primary solution to address the challenges of labor intensity and time consumption. Deep learning is the most effective AI techniques for bioimaging [5]. The identification of leukemia in blood images can be effectively achieved through the implementation of an appropriate deep learning methodology. Deep learning models have demonstrated encouraging outcomes in previous research endeavors. However, the accuracy of these models is susceptible to the conditions of the dataset. Poorly conditioned datasets, characterized by low resolution, low contrast, or blurry images, can lead to inaccurate reports [6].

In our research, we have proposed two convolutional neural network models: one custom model and another based on MobileNetV2 with custom head. Both models demonstrated promising results, with MobileNetV2 performing slightly better. The remainder of the paper is aligned Section 3 delineates the related works, while Section 4 articulates the methods, dataset, and data augmentation. Furthermore, Section 5 elucidates the results and conclusion, respectively.

## II. Related Study

The Significance of Image Processing in Microscopic Blood Images Amidst Rapid Advancements in Digital Technologies Alongside substantial advancements in computational capacity, medical imaging, particularly microscopic images of blood, has become crucial for the early diagnosis of leukemia and its subtypes. Ahmed et al. [7] developed an alternative methodology for noise reduction in microscopic images of blood cells, utilizing the resources available from public ALL-ID Band ASH image bank databases. The CNN model demonstrated a performance accuracy of 88.25% for the ill group and 81.74% for the healthy group. Rehman et al. [8] introduced an efficient convolutional neural network (CNN) model that is trained on bone marrow images to distinguish acute lymphoblastic leukemia (ALL) from microscopic blood samples, achieving an accuracy of 98%. Boldu et al. [9] introduced deep learning models for the classification of leukemia subtypes, with an accuracy of 91.7%. L. Ma et al. [10] introduced a novel architecture that integrates a deep convolutional generative adversarial network (DC-GAN) with a residual neural network (ResNet) for the classification of blood cell pictures, achieving an accuracy of 91.7% on the BCCD dataset. In 2019, Hegde et al. [11] devised an automated method for leukemia identification via an SVM classifier, categorizing white blood cells as normal or abnormal, and attained an accuracy of 92.8%. Dasariraju et al. [12] (2020) developed a random forest method for the detection of acute myeloid leukemia in blood cells. The accuracy attained was 92.99%. They have classified AML into four subtypes with an accuracy of 93.45%. Ramaneswaran et al. [13] (2021) proposed a hybrid model, Inception v3 XGBoost, for the detection of ALL cells from normal cells, achieving an F1 score of 0.986. Sampathila et al. 2022 [14] introduced a bespoke CNN model called ALLNET, which forecasts ALL cells from blood cells with an accuracy of 95.54%. Sulaiman et al. (2023) [15] proposed the ResRandSVM model, which integrates seven pretrained models to identify acute lymphoblastic leukemia (ALL) in blood cell pictures, achieving an accuracy of 0.9000. Ghongade et al. [16] (2023) proposed a deep learning and transfer learning model for the identification of leukemia in blood cell pictures. The models employed were CNN, VGG16, and InceptionV3, with VGG16 attaining the maximum accuracy of 94%.

Our proposed methodology has significantly progressed from previous efforts and integrates an advanced deep learning model.

## III. METHODOLOGY

### A. Dataset and Experiment :

The proposed methodology effectively identifies Acute Lymphoblastic Leukemia (ALL) from blood smear images in the dataset titled "Acute Lymphoblastic Leukemia (ALL) image dataset," utilizing a deep convolutional model. The dataset utilized is derived from actual data of 89 individuals who were suspected of having leukemia, as assessed by competent laboratory personnel at Taleqani Hospital in Tehran, Iran. The collection has a total of 3,256 PBS photos, each with dimensions of 224 × 224 pixels in JPG format. The ALL is categorized by four classes: Benign, Early, Pre, and Pro [17].

We divided the dataset into three segments: Training, Testing, and Validation, with proportions of 80%, 10%, and 10%, respectively. Figure 1(a). Displays the dataset's state before to any processing. The training data is uneven across the four classes, which may lead to overfitting issues. To address this, we have employed the Synthetic Minority Oversampling Technique (SMOTE). SMOTE is an oversampling algorithm that equilibrates data by generating synthetic instances. Figure 1(b) illustrates the balanced dataset.

The dataset has been normalized to facilitate more rapid convergence. Label encoding is employed to transform the labels into a purely numerical format. Shuffling is employed to address the issue of overfitting to sequence. Figure 2 presents the input data along with their respective labels.

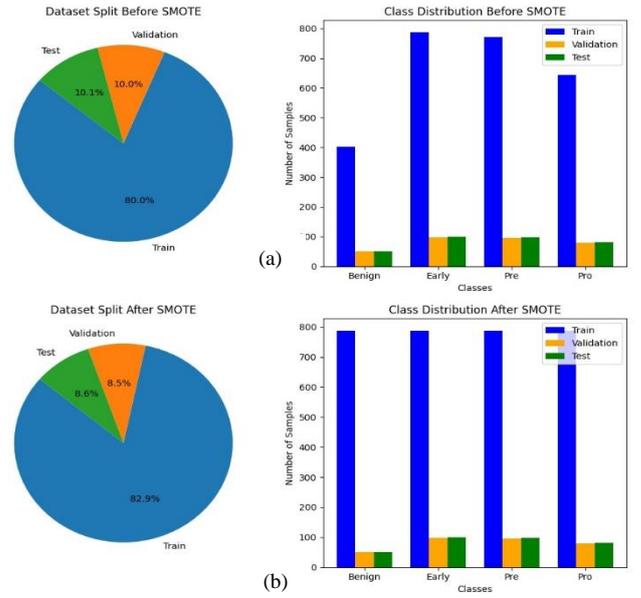

Fig. 1. Dataset split and applying SMOTE (a) Before (b) After

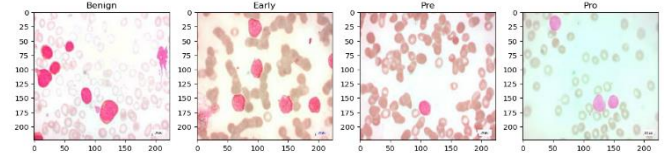

Fig. 2. Sample Blood Smear Images of the Acute Lymphoblastic Leukemia (ALL) dataset for each class

### B. Training Models :

#### 1) Deep Transfer Learning Model:

*a) MobileNetV2:* MobileNetV2 is a lightweight convolutional neural network architecture characterized by accelerated convergence and reduced error rates [18]. The classification head has been modified for the proposed MobileNetV2 architecture. The revised head comprises a singular global average pooling layer, accompanied by fully connected dense layers. The output layer employs the softmax function as its activation mechanism. The Rectified Linear Unit (ReLU) is one of the most widely utilized activation functions in deep learning models, particularly in the domain of computer vision. This phenomenon occurs because the ReLU activation function selects the maximum of the input values, thereby disregarding negative values by assigning them a value of zero. The role of the ReLU function,

$$relu(x) = max(0, x) \quad (1)$$

Softmax activation function is used to calculate probabilities of a multi-class detection system. The mathematical representation of the softmax.

$$softmax(x) = \frac{e^{x_i}}{\sum_j e^{x_j}} \quad (2)$$

Softmax is extensively utilized for multi-class detection. This serves as the rationale for its utilization in our classification layer.

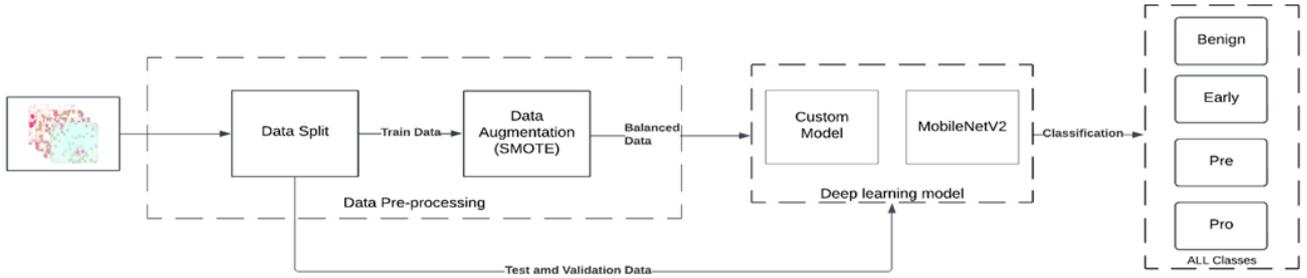

Fig. 3. Block diagram of multiclass classification using the proposed framework

Categorical Crossentropy: This metric is employed to assess the discrepancy between the predicted class and the actual class. The primary purpose of employing this function is backpropagation. The derivative of the cross-entropy function exhibits a greater magnitude for poor predictions, thereby facilitating the process of backpropagation. The mathematical representation of Crossentropy,

$$L(y, \hat{y}) = -\Sigma_{i=1}^{c} y_i \log(\hat{y}) \qquad (3)$$

*2) Custom CNN Model:* Figure 3 illustrates the model architecture in a three-dimensional format. There exist three convolutional layers utilizing the ReLU activation function, comprising 32, 64, and 128 neurons as outputs, along with their respective max pooling layers. The subsequent flattening layer generates input data for the fully connected layer comprising 128 neurons, from which 50% of the data is excluded through the application of a dropout layer. Subsequently, a dense layer is employed as the output, utilizing the softmax activation function, which comprises four output neurons corresponding to four distinct classes. The input dimensions are 224 by 224 pixels and are represented in RGB format.

*C. Hyperparameters :*

The parameters that are established prior to training significantly influence the training process. The parameters such as epochs, batch size, learning rate, optimizer, and dropout rate are classified as hyperparameters. Table I presents the hyperparameters associated with both models. These values demonstrate the optimal outcomes for the subsequent models.

IV. RESULT AND DISCUSSION

The proposed methodology has demonstrated significant results for both models. The Receiver Operating Characteristic (ROC) curve, the Confusion Matrix, and classification reports, including the F1 score, precision, and recall, are computed for both models. Figures 5 and 6 illustrate the ROC curve and the confusion matrix, respectively. Table III presents a comparative analysis of the models we have proposed in relation to those that have been previously introduced.

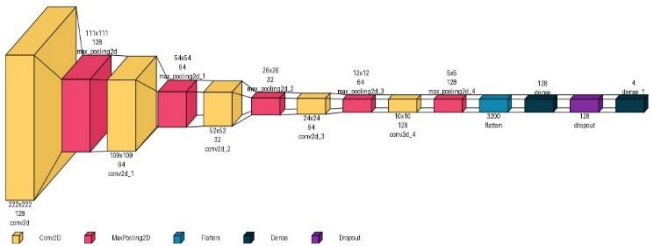

Fig. 4. Architecture of Custom CNN

Table I
MODELS PARAMETERS FOR INPUT AND CLASSIFICATION STAGE

| Parameters | Approach |
|---|---|
| Input Size | $224 \times 244$ |
| No. of Epochs | 150 |
| Batch size | 16 |
| Activation | Softmax |
| Optimizer | Adam |
| Learning Rate | 0.0001 |

Analysis of the ROC curve reveals that the custom model exhibits the lowest Area Under Curve (AUC) value of 0.9807, whereas MobileNetV2 demonstrates a notably higher AUC of 0.9990, representing the lowest among the evaluated models. Upon examining the Confusion Matrix presented in Figure 6, it is evident that the diagonal values, which signify the accurately predicted outcomes, indicate that the Confusion Matrix for MobileNetV2 demonstrates superior performance in this context.

From Table III, it can be concluded that the proposed methodology exceeds recent research in the relevant area. The proposed methodology demonstrates promising results, characterized by enhanced convergence speed and reduced weight.

V. CONCLUSION

The identification of leukemia represents a significant challenge within the diagnostic process of cancer. The timely identification of leukemia can significantly influence the diagnostic process. Consequently, numerous studies are undertaken in the domain of vision computation to identify leukemia. However, the majority of them are deficient in empirical data. The proposed methodology primarily involves the classification of white blood cells to differentiate between benign cells and those associated with Acute Lymphoblastic Leukemia (ALL). This is achieved through the utilization of a custom model and a custom head based on MobileNetV2, employing pretrained weights from ImageNet. The achievement of 99.69 percent accuracy was realized through the utilization of MobilenetV2, while a training accuracy of 98.6 percent was attained with the Custom model. The accuracy attained by these models is indeed remarkable and holds significant promise. The system identifies the presence of leukemia and, if detected, classifies it into three distinct categories: Early, Pre, and Pro. Consequently, the results of our experiments demonstrate greater reliability in real-world scenarios. The predictions provided by the model are firmly established, as the labels for the dataset are assigned by expert laboratory personnel.

Table II
COMPARISON OF THE PROPPOSED MODELS WITH OTHER DEEP LEARNING APPROACHES

| References | No. of Classes | Used Models | Precision | Recall | F1 Score | Accuracy |
|---|---|---|---|---|---|---|
| [8] | 04 | Custom CNN | - | - | - | 97.78% |
| [13] | 02 | InceptionV3 XGBoost | 0.979 | 0.979 | 0.986 | 97.90% |
| [14] | 02 | ALLNET | 0.96 | 0.9591 | 0.9543 | 95.45% |
| [15] | 02 | ResRandSVM | 0.902 | 0.957 | 0.929 | 90.00% |
| [16] | 04 | CNN | 0.79 | - | - | 84% |
| | | VGG16 | 0.86 | - | - | 94% |
| | | InceptionV3 | 0.81 | - | - | 89% |
| Proposed Model | 04 | Custom CNN | 0.98 | 0.99 | 0.98 | 98.6% |
| | | MobileNetV2 | 0.99 | 0.99 | 0.99 | 99.69 |

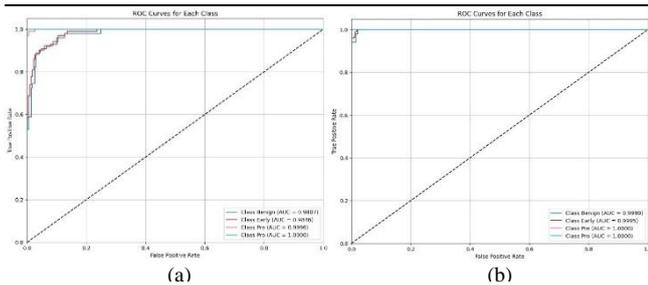

Fig. 5. ROC curve of (a) Custom Model and (b) MobileNetV2

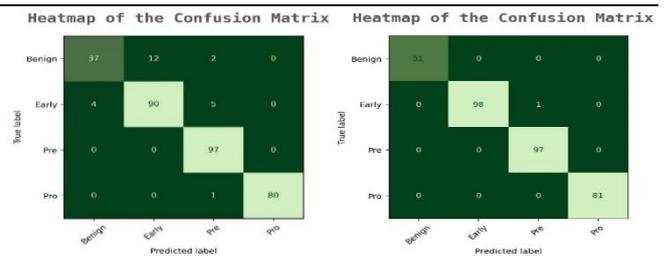

Fig. 6 Confusion matrix of (a) Custom Model and (b) MobileNetV2